\documentclass{article}
\usepackage[a4paper,margin=1in]{geometry}
\usepackage{graphicx}
\usepackage{caption}
\usepackage{amsmath}
\usepackage{amssymb}
\usepackage{float}
\usepackage{subcaption}
\usepackage{natbib}

\begin{document}

\title{Enhancing diffusion models with Gaussianization preprocessing}
\author{
Li Cunzhi$^{1}$, Louis Kang$^{1,2}$, Hideaki Shimazaki$^{1}$ \\
{\small $^{1}$Graduate School of Informatics, Kyoto University, Kyoto, Japan} \\
{\small $^{2}$Neural Circuits and Computations Unit, RIKEN Center for Brain Science, Saitama, Japan}
}

\date{}

\maketitle

\begin{abstract}
Diffusion models are a class of generative models that have demonstrated remarkable success in tasks such as image generation. However, one of the bottlenecks of these models is slow sampling due to the delay before the onset of trajectory bifurcation, at which point substantial reconstruction begins. This issue degrades generation quality, especially in the early stages. Our primary objective is to mitigate bifurcation-related issues by preprocessing the training data to enhance reconstruction quality, particularly for small-scale network architectures. Specifically, we propose applying Gaussianization preprocessing to the training data to make the target distribution more closely resemble an independent Gaussian distribution, which serves as the initial density of the reconstruction process. This preprocessing step simplifies the model's task of learning the target distribution, thereby improving generation quality even in the early stages of reconstruction with small networks. The proposed method is, in principle, applicable to a broad range of generative tasks, enabling more stable and efficient sampling processes.
\end{abstract}


\section{Introduction}

Diffusion models \citep{sohl2015deep,ho2020denoising,song2020score} have emerged as one of the most powerful classes of generative models for high-dimensional data, achieving state-of-the-art performance in image synthesis \citep{dhariwal2021diffusion,rombach2022high} and other tasks such as action generation in robotic or protein design \citep{watson2023novo,chi2025diffusion}. However, sampling from these models is typically slow: many reverse-time steps are required to transform an initial Gaussian sample into a high-quality sample in data space \citep{ho2020denoising,song2020score}. This computational cost is especially problematic, and it restricts the practical deployment of diffusion models in real-time or resource-constrained settings \citep{salimans2022progressive,lu2022dpm}.

Recent theoretical and empirical studies suggest that this inefficiency is closely related to a dynamical phase transition (bifurcation) that occurs during the reverse process \citep{raya2024spontaneous,biroli2024dynamical,ambrogioni2025statistical}. In the early reverse steps, the trajectories stay near a stable fixed point whose distribution is close to the initial independent Gaussian, and little structure is present in the samples. At a critical time, this fixed point loses stability and trajectories bifurcate toward clusters corresponding to memorized patterns in the data. It has been shown that skipping the initial pre-transition period does not degrade sample quality, and that starting the reverse process just before the bifurcation point (“Gaussian late start”) can substantially reduce sampling cost \citep{raya2024spontaneous}. In practice, however, the transition time depends on the data distribution and the trained network, and is difficult to predict for realistic datasets such as natural images \citep{biroli2024dynamical}.

A key factor behind this bifurcation behavior is the mismatch between the independent standard Gaussian distribution used as the starting point of inference and the non-Gaussian distribution of real data in pixel or feature space. The bifurcation is likely to happen because the distributions of the memorized patterns in the data space are typically clustered, which significantly differs from the initial Gaussian distribution. The reverse trajectories must travel a long distance in distribution space before reaching high-density regions corresponding to the data, leading to an extended pre-transition regime. If the target distribution were closer to an independent Gaussian, the reverse dynamics would not need such a long approach phase, and high-quality samples could be obtained in fewer steps. This observation suggests an alternative strategy: instead of only modifying the dynamics or the time schedule of the diffusion process, we can preprocess the training data so that its distribution more closely resembles the independent Gaussian used as the initial density. Provided that this preprocessing is invertible and avoids information loss, we can train diffusion models in the transformed space and map generated samples back to the original data space.

In this work, we propose such a preprocessing approach based on Gaussianization \citep{chen2000gaussianization,laparra2011iterative}. Our approach combines Independent Component Analysis (ICA) with one-dimensional Gaussianization of each component to transform the data distribution toward an independent standard Gaussian. Concretely, we use an iterative Gaussianization scheme in which alternating applications of ICA and Gaussianization progressively eliminate dependencies and non-Gaussian structure \citep{laparra2011iterative}. Because the transformed data are already close to the initial noise distribution of the diffusion process, the reverse trajectories become simpler and the model can reconstruct high-quality samples in fewer steps, even for small networks. The inverse transformation allows us to recover samples in the original data space without relying on lossy compression or a separately trained autoencoder.

We validate this approach in controlled experiments on synthetic data generated from Gaussian mixture models (GMMs). Comparing diffusion models trained on raw data with those trained on Gaussianized data, we show that Gaussianization preprocessing leads to faster convergence in inference time and modest acceleration of training, while maintaining or improving alignment with the true data distribution as measured by average log-likelihood. In particular, the Gaussianized pipeline attains high log-likelihood values within the first tens of reverse steps and exhibits smoother, more stable reconstruction trajectories, whereas the baseline trained on raw data typically requires many more steps and passes through a bifurcation regime. These findings provide a link between dynamical phase transitions in diffusion models and the geometry of the data distribution \citep{biroli2024dynamical,ambrogioni2025statistical} and highlight data-space transformations such as Gaussianization as a complementary axis for improving the efficiency and stability of diffusion-based generative modeling.

\subsection{Diffusion Models}

Denoising Diffusion Probabilistic Models (DDPMs) are a class of generative models that have demonstrated remarkable success in tasks such as image generation, natural language processing, and video synthesis. These models rely on a two-stage process: {forward diffusion} and {reverse denoising}, rooted in principles from thermodynamics and stochastic processes.

In the {forward diffusion process}, noise is iteratively added to the data over a series of time steps, progressively transforming the data distribution into a simple Gaussian distribution. This process is described by a Markov chain:
\begin{equation}
q(x_t \mid x_{t-1}) = \mathcal{N}(x_t; \sqrt{1 - \beta_t} x_{t-1}, \beta_t I),
\end{equation}
where $\beta_t$ is the noise scheduling parameter that controls the rate of noise addition at each time step $t$. Typically, $\beta_t$ is designed to gradually increase over time to ensure a smooth transition of the data distribution from its original state to a simple Gaussian distribution. Linear noise scheduling is a widely used strategy in diffusion models, where the noise level \(\beta_t\) increases linearly over the diffusion steps. This straightforward approach effectively balances the trade-off between data distortion and model stability, leading to high-quality generated samples. The term $\sqrt{1 - \beta_t}$ acts as a scaling factor to balance the contribution of the signal and noise at each step. The direct relationship between $x_t$ and $x_0$ can be expressed as:
\begin{equation}
x_t = \sqrt{\bar{\alpha}_t} x_0 + \sqrt{1 - \bar{\alpha}_t} \epsilon, \quad \epsilon \sim \mathcal{N}(0, I),
\label{eq:x_t_from_x_0}
\end{equation}
where $\bar{\alpha}_t = \prod_{s=1}^t (1 - \beta_s)$ represents the cumulative noise schedule.

The {reverse process} aims to iteratively denoise the data, reconstructing it step by step through a neural network that predicts the added noise. This process is parameterized as:
\begin{equation}
p_\theta(x_{t-1} \mid x_t) = \mathcal{N}(x_{t-1}; \mu_\theta(x_t, t), \Sigma_\theta(x_t, t)),
\end{equation}
where $\mu_\theta(x_t, t)$ and $\Sigma_\theta(x_t, t)$ represent the neural network’s parameterized predictions for the mean and variance, respectively. To simplify the optimization, the variance $\Sigma_\theta(x_t, t)$ is often fixed as a predefined function (e.g., $\Sigma_\theta(x_t, t) = \beta_t$), and the focus is placed on optimizing the mean $\mu_\theta(x_t, t)$. The true posterior $q(x_{t-1} \mid x_t, x_0)$ is given by:
\begin{equation}
q(x_{t-1} \mid x_t, x_0) = \mathcal{N}(x_{t-1}; \tilde{\mu}_t(x_t, x_0), \tilde{\beta}_t I),
\end{equation}
where:
\begin{equation}
\tilde{\mu}_t(x_t, x_0) = \frac{1}{\sqrt{\alpha_t}} \left(x_t - \frac{\beta_t}{\sqrt{1 - \bar{\alpha}_t}} \epsilon \right),
\end{equation}
\begin{equation}
\tilde{\beta}_t = \frac{1 - \bar{\alpha}_{t-1}}{1 - \bar{\alpha}_t} \beta_t.
\end{equation}

The neural network $\epsilon_\theta(x_t, t)$ is trained to predict the noise $\epsilon$ added during the forward process. The optimization objective simplifies to:
\begin{equation}
\mathcal{L}_{\text{simple}} = \mathbb{E}_{x_0, \epsilon, t} \left[ \| \epsilon - \epsilon_\theta(x_t, t) \|^2 \right],
\end{equation}
where $x_t$ is given by Eq.~\ref{eq:x_t_from_x_0}. This loss function directly minimizes the mean squared error (MSE) between the true noise $\epsilon$ and the predicted noise $\epsilon_\theta(x_t, t)$, ensuring that the model learns to denoise effectively and reconstruct the original data distribution.

As described by Ho et al.~\citep{ho2020denoising}, DDPMs are considered a leading technique in generative modeling due to their state-of-the-art performance in generating realistic and diverse samples with minimal mode collapse.

\subsection{Previous works on reducing computational costs of diffusion model}

Diffusion models (DMs) have achieved state-of-the-art performance in image synthesis and other generative tasks. However, their computational requirements for training and sampling remain a significant challenge. To address these challenges, several preprocessing strategies have been proposed, including DiffuseVAE \citep{pandey2022diffusevae} and Latent Diffusion Models (LDMs) \citep{rombach2022high}. These methods leverage latent space representations to improve the efficiency and quality of diffusion processes.

\subsubsection*{DiffuseVAE: Leveraging VAE for Preprocessing}

DiffuseVAE \citep{pandey2022diffusevae} introduces a two-stage framework that combines the strengths of Variational Autoencoders (VAEs) and diffusion models. In essence, in the first stage, a VAE is used to map the original data $x$ into a low-dimensional latent space $z$, generating blurry samples that capture the primary structure of the data:
\[
q_\psi(z \mid x), \quad p_\phi(x\mid z), 
\]
where \( q_\psi \) is the VAE encoder and \( p_\phi \) is the decoder. These blurry samples serve as the starting point for the second stage, where a diffusion model refines the samples to enhance their visual fidelity. This refinement process is governed by the conditional distribution:
\[
x_0 = \text{DDPM}_\theta(x, z),
\]
allowing the diffusion model to add fine details while preserving the global structure encoded in the latent space.

DiffuseVAE offers several benefits. First, the low-dimensional latent space provides direct control over the major structure of the generated samples, enabling controllable synthesis. Second, by focusing the diffusion process on refining blurry samples, DiffuseVAE reduces the number of required sampling steps, improving efficiency.  Additionally, the method generalizes well to multi-modal generation tasks, such as class-conditional synthesis. However, the effectiveness of DiffuseVAE heavily depends on the quality of the VAE's encoder-decoder pair, as poor reconstruction quality in the first stage can hinder the diffusion model's performance.

\subsubsection*{Latent Diffusion Models: Efficient Training in Latent Space}

Latent Diffusion Models (LDMs) \citep{rombach2022high} further optimize diffusion processes by operating directly in the latent space of a pretrained autoencoder. Instead of applying diffusion to the high-dimensional pixel space, LDMs compress the data into a perceptually equivalent, lower-dimensional representation:
\[
z = E(x), \quad x = D(z),
\]
where \( E \) and \( D \) are the encoder and decoder of the autoencoder, respectively. Once compressed, the diffusion model operates in the latent space, modeling the distribution of \( z \) through:
\[
p_\theta(z_{t-1} \mid z_t) = \mathcal{N}(z_{t-1}; \mu_\theta(z_t, t), \Sigma_\theta(z_t, t)).
\]
After the latent space sampling, the decoder reconstructs the samples back into the pixel space:
\[
x_{\text{gen}} = D(z_{\text{gen}}).
\]

LDMs significantly reduce computational complexity, as the dimensionality of the latent space is much smaller than that of the original pixel space. This reduction allows for faster training and inference while maintaining high visual fidelity. Additionally, by incorporating cross-attention layers, LDMs enable flexible conditioning mechanisms, such as text or bounding box inputs, making them suitable for high-resolution image synthesis tasks. The method achieves state-of-the-art performance on various tasks, including unconditional image generation, super-resolution, and inpainting. However, like DiffuseVAE, LDMs rely on the quality of the autoencoder; if the latent representation loses essential semantic information, the diffusion model's performance may be compromised.

Both DiffuseVAE and LDMs highlight the importance of preprocessing in diffusion models. DiffuseVAE focuses on leveraging the semantic control of VAEs to refine blurry samples, whereas LDMs exploit a pretrained autoencoder to reduce computational demands by compressing the data into a latent space. Despite their advantages, these methods share a reliance on pretrained models, which introduces additional training complexity and potential performance bottlenecks.

\section{Method}

\subsection{Gaussianization Method}

\begin{figure}[t]
    \centering
    \subfloat[Original Data]{
        \includegraphics[width=0.22\textwidth]{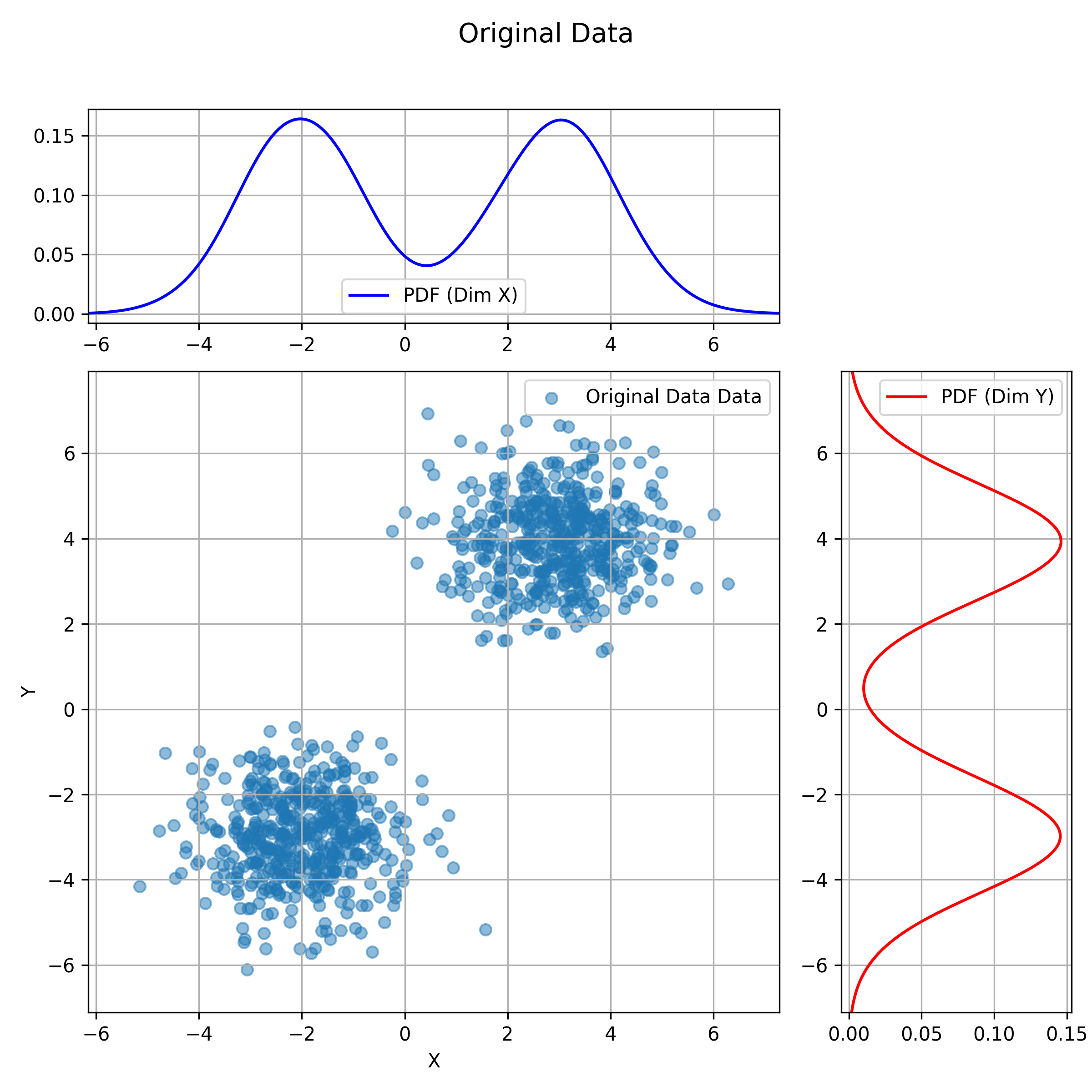}
    }
    \subfloat[ICA Data]{
        \includegraphics[width=0.22\textwidth]
        {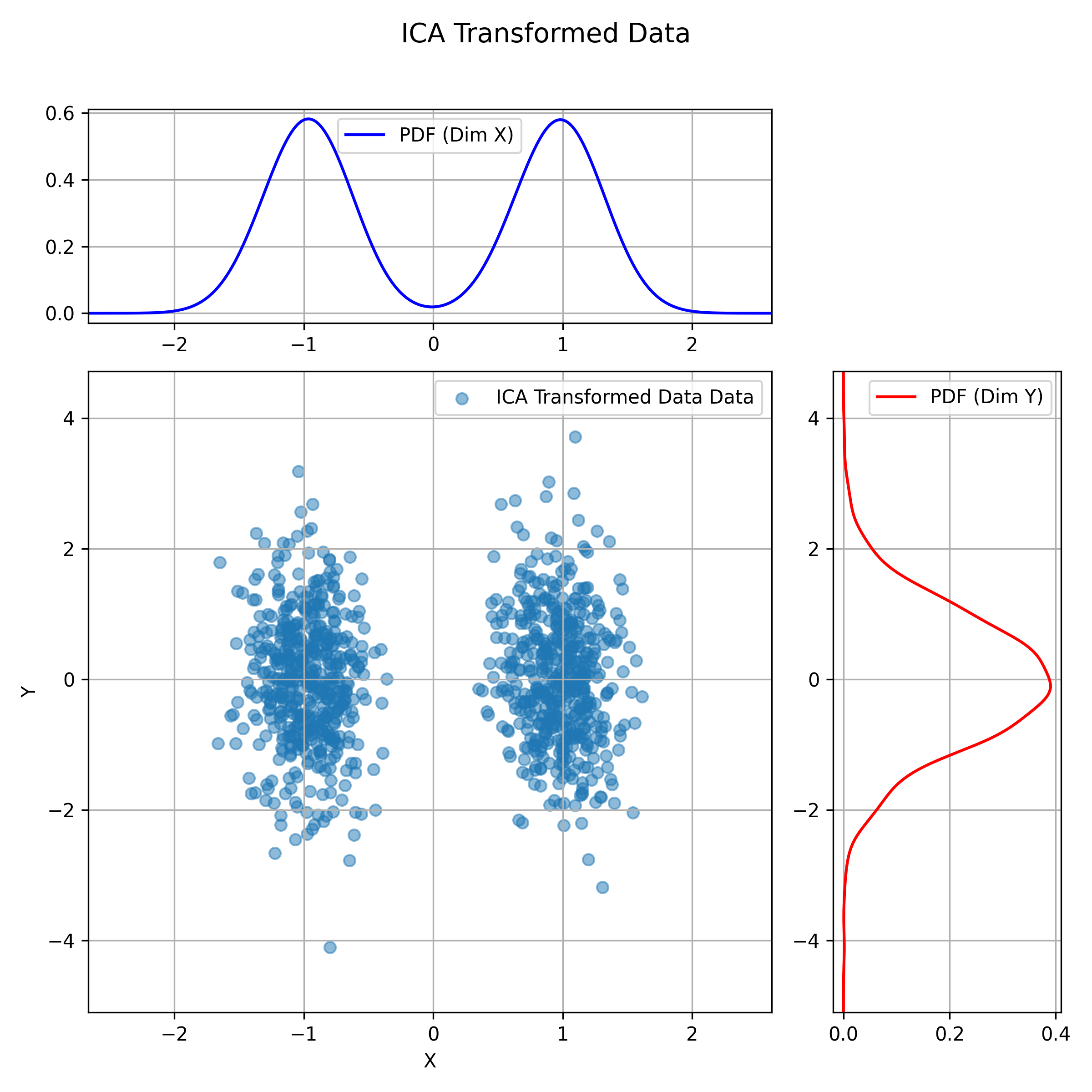}
    }
    \subfloat[Gaussianization]{
        \includegraphics[width=0.22\textwidth]{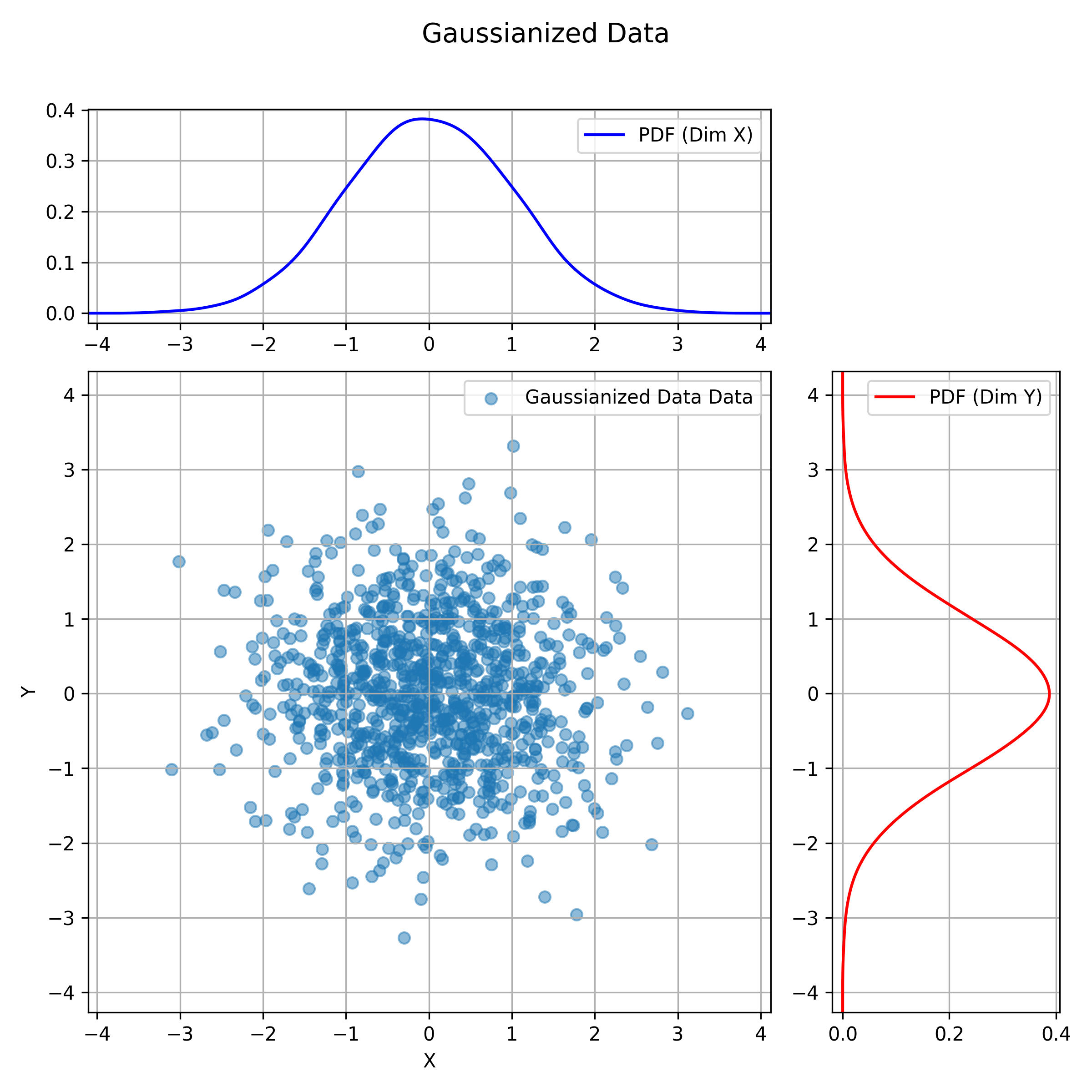}
    }
    \subfloat[Reconstruction]{
        \includegraphics[width=0.22\textwidth]{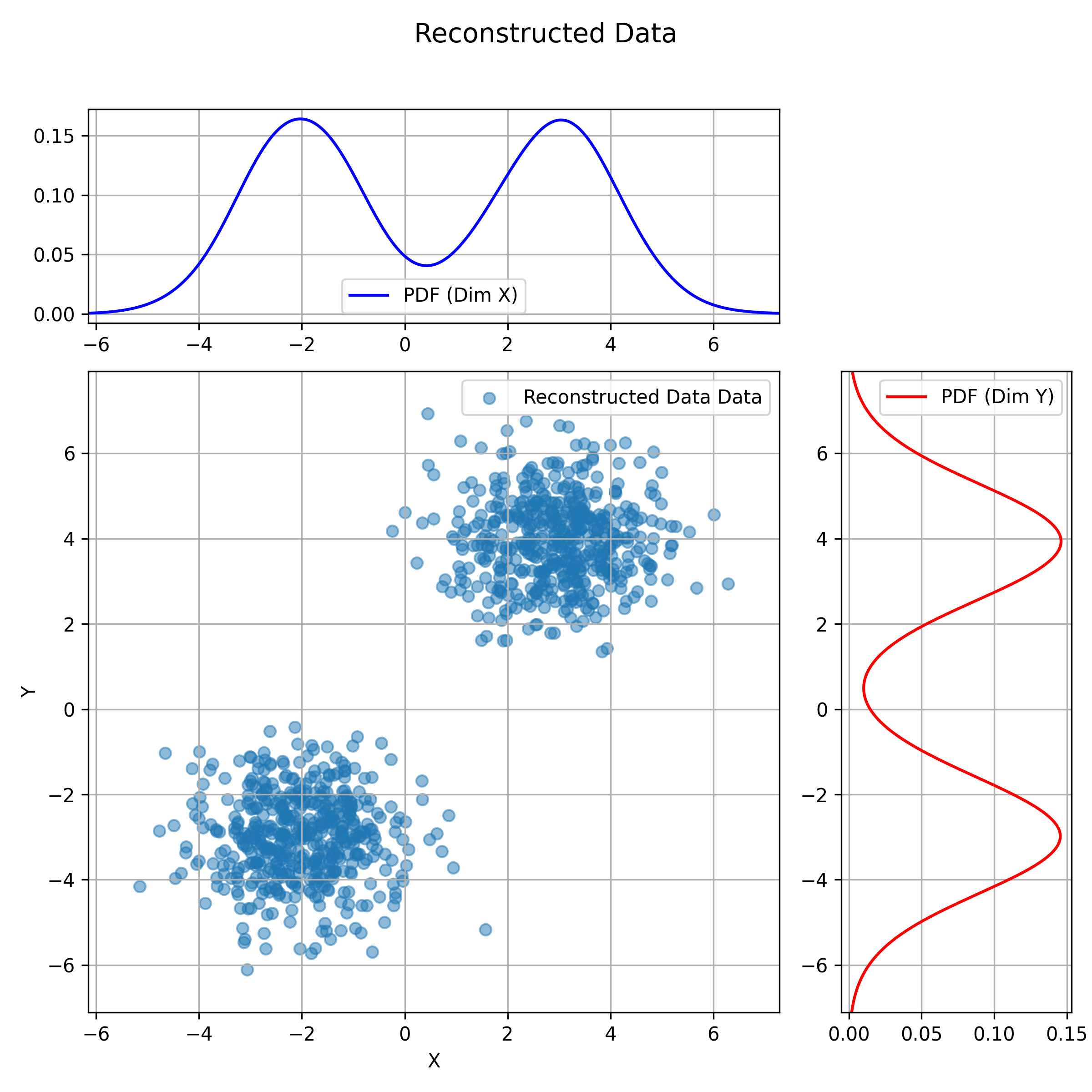}
    }
    \caption{The data transformation process. From left to right: Original data distribution, ICA transformed data distribution, Gaussianized data distribution, and reconstructed data distribution. Each step visualizes the evolution of the data through the proposed pipeline.}
    \label{fig:distributions}
\end{figure}

\begin{figure}[t]
    \centering
    \includegraphics[width=0.8\textwidth]{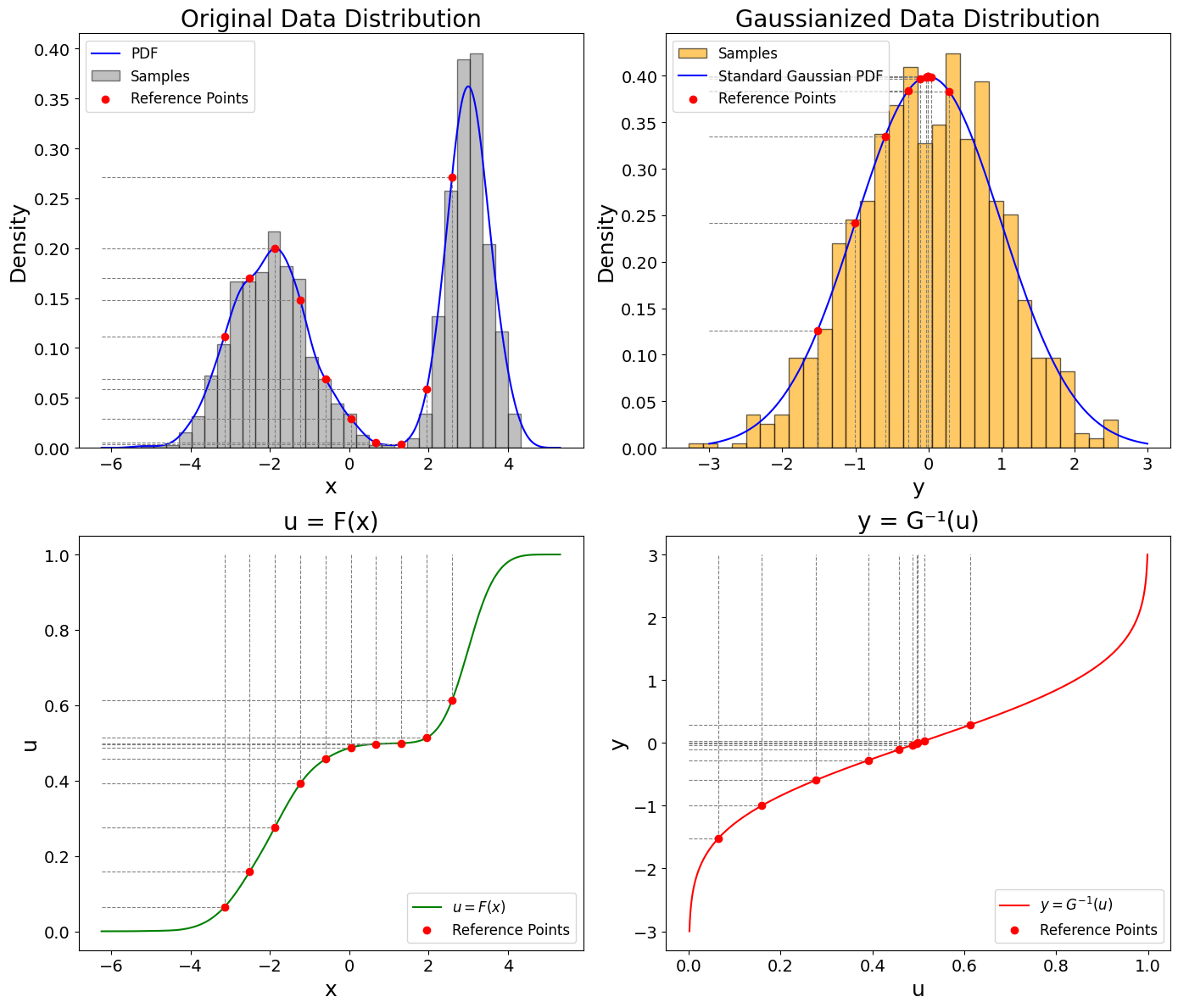}
    \caption{Illustration of the one-dimensional Gaussianization process.
    Top Left: Original bimodal distribution. Bottom Left: \( u = F(x) \). Bottom Right: \( y = G^{-1}(u) \). Top Right: Gaussianized distribution.}
    \label{fig:gaussianization}
\end{figure}

Our method consists of two steps (Fig.~\ref{fig:gaussianization}). The first step aims to make the distribution independent using ICA, which leverages non-Gaussianity as a measure of independence. By iteratively maximizing the non-Gaussianity of projected data, ICA identifies the most non-Gaussian directions, which are likely to correspond to independent components according to the central limit theorem \citep{hyvarinen2000independent}. The second step transforms the marginal distributions into a Gaussian form using a Gaussianization process~\citep{chen2000gaussianization}, which applies a sequence of transformations to map the data onto a standard Gaussian distribution.


\subsubsection*{Extracting Independent Components using Independent Component Analysis (ICA)}

To preprocess the data, we first extract statistically independent components using Independent Component Analysis (ICA). Let \( x \in \mathbb{R}^d \) denote the original data vector. ICA transforms \( x \) into independent components \( z \in \mathbb{R}^d \) using the ICA transformation function:
\begin{equation}
z = \text{ICA}(x),
\end{equation}
where \( z = [z_1, z_2, \dots, z_d]^\top \) and each \( z_i \) represents the \( i \)-th independent component, statistically independent from all other components.

\subsubsection*{Gaussianizing Each Independent Component}

After obtaining the independent components \( z \), we apply Gaussianization to transform each \( z_i \) into a standard normal distribution. This process consists of three steps:

\paragraph{Step 1: Estimating the Probability Density Function (PDF)}

For each independent component \( z_i \), we estimate its probability density function \( p_i(z) \) using Kernel Density Estimation (KDE):
\begin{equation}
\hat{p}_i(z) \approx \frac{1}{nh} \sum_{j=1}^n K\left(\frac{z - z_{ij}}{h}\right),
\end{equation}
where \( K \) is a kernel function and \( h \) is the bandwidth parameter. Here, \( z_{ij} \) represents the \( j \)-th observed sample of the \( i \)-th independent component, and \( n \) denotes the total number of samples.

A smaller bandwidth \( h \) enables the kernel density estimation to capture finer details of the data distribution, allowing it to approximate the standard Gaussian distribution more closely. However, excessively small bandwidths may lead to overfitting, resulting in noisy density estimates \citep{silverman2018density}. This, in turn, makes the cumulative distribution function (CDF) less smooth and the inverse mapping in the reconstruction process prone to errors, ultimately reducing the accuracy of the restored data. While here we selected \( h \) manually, one may use a data-driven optimization method of the kernel bandwidth \citep{shimazaki2010kernel}.

\paragraph{Step 2: Mapping to a Uniform Distribution via the Cumulative Distribution Function (CDF)}
The cumulative distribution function (CDF) of \( z_{ij} \), denoted as \( F_i(z) \), is computed by integrating its PDF:
\begin{equation}
F_i(z) = \int_{-\infty}^{z} \hat{p}_i(t) \, dt.
\end{equation}
Using \( F_i(z) \), we map \( z_{ij} \) to a uniform distribution on the interval \([0, 1]\):
\begin{equation}
u_{ij} = F_i(z_{ij}).
\end{equation}
This transformation is known as the probability integral transform, which states that if a random variable \( z_{ij} \) has a continuous cumulative distribution function \( F_i(z) \), then the transformed variable \( u_{ij} = F(z_{ij}) \) follows a uniform distribution on \([0,1]\). 

\paragraph{Step 3: Transforming to a Standard Normal Distribution via the Inverse CDF}

To Gaussianize \( z_{ij} \), we apply the inverse CDF of the standard normal distribution \( {G}^{-1} \):
\begin{equation}
\mathcal{Z}_{ij} = {G}^{-1}(u_{ij}).
\end{equation}
Here, \( \mathcal{Z}_{ij} \) represents the Gaussianized samples of the $i$-th independent component.

\subsection{Inverse Gaussianization method}

\subsubsection*{Recovering the Original Independent Components}

To reconstruct the original independent components from the Gaussianized data, we reverse the Gaussianization process:

\begin{itemize}
    \item \textbf{Step 1: Mapping back to the uniform distribution}
For each Gaussianized independent component \( \mathcal{Z}_{ij} \), we compute its CDF under the standard normal distribution:
\begin{equation}
u_{ij} = {G}(\mathcal{Z}_{ij}).
\end{equation}
    \item \textbf{Step 2: Mapping back to the original distribution}
Using the inverse of the previously computed CDF \( F_i^{-1}(u) \), we map \( u_{ij}\) back to the original independent component:
\begin{equation}
z_{ij}= F^{-1}(u_{ij}).
\end{equation}
\end{itemize}

\subsubsection*{Recovering the Original Data using Inverse ICA}

Finally, the original data vector \( x \) is reconstructed by applying the inverse ICA transformation:
\begin{equation}
x = \text{ICA}^{-1}(z).
\end{equation}

This completes the Gaussianization method, which transforms data into a Gaussianized form and subsequently reconstructs the original data.

\subsection{Iterative Gaussianization Method}

While ICA is effective at detecting and transforming the directions with the greatest deviation from Gaussianity, real-world data may still contain residual non-Gaussian behaviors along other axes after applying Gaussianization. These residual characteristics reflect the fact that complex and high-dimensional data distributions are often not completely linearly independent, even after one round of ICA. Furthermore, ICA is inherently a non-orthogonal transformation, meaning that transforming the distribution along one independent component (IC) axis may interfere with Gaussianization along other axes.

To address these challenges, we perform an iterative Gaussianization method that performs multiple rounds of the Gaussianization method introduced above \citep{laparra2011iterative}. To briefly explain again, each iteration consists of the two steps of Independent Component Analysis (ICA) and Gaussianization. In the first step, ICA identifies the most non-Gaussian directions within the data. In the second step, the marginal distributions of the identified independent components are Gaussianized. To achieve this, the method leverages kernel density estimation (KDE), a non-parametric density estimation technique.

By repeatedly applying the ICA and Gaussianization steps, this method progressively eliminates correlations and non-Gaussian characteristics within the data, ultimately transforming the distribution into an independent Gaussian form. 

\subsubsection*{Forward Gaussianization Process}
Let $x^{(0)} \in \mathbb{R}^d$ denote the original data. The iterative Gaussianization process starts from $x^{(0)}$ and produces a Gaussianized representation $\mathcal{Z}^{(K)}$ after $K$ iterations. In each iteration $k$, the following steps are performed:

\begin{itemize}
    \item 
\textbf{Step 1: Independent Component Analysis (ICA)}
The data $x^{(k)}$ is transformed into components $z^{(k)}$ using ICA to maximize their statistical independence:
\begin{equation}
    z^{(k)} = \text{ICA}(x^{(k)}),
\end{equation}
where $z^{(k)} = [z^{(k)}_1, z^{(k)}_2, \dots, z^{(k)}_d]^\top$ represents the independent components.
    \item 
\textbf{Step 2: Gaussianization of Independent Components}
Each component \( z^{(k)}_i \) is Gaussianized to produce \( \mathcal{Z}^{(k)}_i \), using the following methods.

\textbf{Kernel Density Estimation (KDE):}  
The probability density function (PDF) \( \hat{p}^{(k)}_i(z) \) is approximated using KDE:  
\begin{equation}
    \hat{p}^{(k)}_i(z) = \frac{1}{nh} \sum_{j=1}^n K\left( \frac{z - z^{(k)}_{ij}}{h} \right),
\end{equation}
where \( K \) is a kernel function and \( h \) is the bandwidth parameter. Here, \( z^{(k)}_{ij} \) represents the \( j \)-th sample of the $i$-th independent component at iteration \( k \), and \( n \) is the total number of samples.

\textbf{Cumulative Distribution Function (CDF):}  
The CDF \( F^{(k)}_i(z) \) is computed by integrating the PDF:  
\begin{equation}
    F^{(k)}_i(z) = \int_{-\infty}^{z} \hat{p}^{(k)}_i(t) \, dt.
\end{equation}  

\textbf{Gaussian Mapping:}  
The standard Gaussian inverse CDF \( {G}^{-1} \) is used to transform the data:  
\begin{equation}
    \mathcal{Z}^{(k)}_{ij} = {G}^{-1}(F^{(k)}_i(z^{(k)}_{ij})),
\end{equation}
where \( {G}^{-1} \) maps the uniform distribution derived from \( F(z^{(k)}_{ij}) \) to a standard Gaussian distribution.  

    \item 
\textbf{Step 3: Initialization of the next iteration:} 
At the end of each iteration, the Gaussianized data \( \mathcal{Z}^{(k)} \) becomes the input for the next iteration as
\begin{equation}
    x^{(k+1)} = \mathcal{Z}^{(k)}.
\end{equation}  
\end{itemize}

\subsubsection*{Inverse Reconstruction Process}
To recover the original data \( x^{(0)} \) from the Gaussianized representation \( \mathcal{Z}^{(K)} \), the reverse process iterates from the last iteration \( K \) back to the initial step. For each iteration \( k \), the following steps are performed:  

\begin{itemize}
    \item 
\textbf{Step 1: Inverse Gaussianization}

The inverse of the previously computed CDF maps the Gaussianized data back to its ICA components:
\begin{equation}
    z^{(k)}_{ij} = (F^{(k)}_i)^{-1}({G}(\mathcal{Z}^{(k)}_{ij})).
\end{equation}  
    \item
\textbf{Step 2: Inverse ICA Transformation}
The partially reconstructed data at step \( k \) is reconstructed using the inverse ICA transformation:  
\begin{equation}
    x^{(k)} = \text{ICA}^{-1}(z^{(k)}).
\end{equation}  
\end{itemize}

The iterative reconstruction process follows a stepwise approach, where at each iteration \( k \), the partially reconstructed data is updated by reversing the transformations applied in the forward process. Specifically, the relationship between the transformed and original data at step \( k \) is given by:
\begin{equation}
    \mathcal{Z}^{(k-1)} = x^{(k)}.
\end{equation}

Each step iteratively refines the reconstruction, ensuring that the final recovered data \( x^{(0)} \) closely matches the original dataset before transformation. This iterative Gaussianization method ensures that complex data distributions can be effectively simplified into Gaussian representations while maintaining the ability to accurately reconstruct the original data.

\begin{figure}[t!]
    \centering
    \includegraphics[width=0.80\linewidth]
    {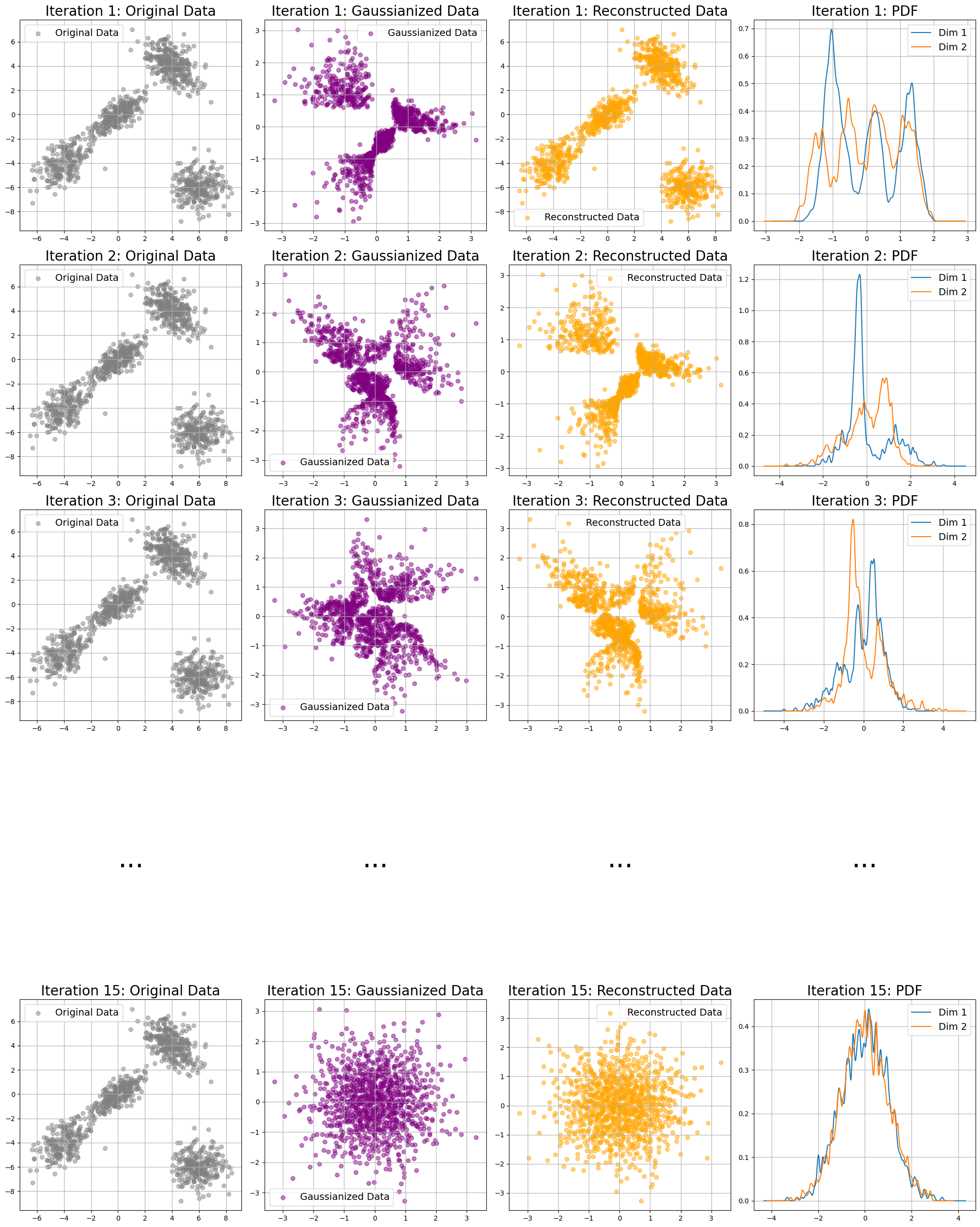}
    \caption{Iterative Gaussianization Process. The figure demonstrates the transformation of a two-dimensional dataset through the Gaussianization pipeline across multiple iterations. Each row corresponds to a specific iteration, showing the original data distribution $x^{(0)}$, the Gaussianized data $\mathcal{Z}^{(k)}$, the partially reconstructed data $x^{(k)} = \mathcal{Z}^{(k-1)}$, and the corresponding probability density functions (PDFs) $\hat{p}^{(k)}_i(z)$ for both dimensions. The iterative process ensures the reconstructed data aligns closely with the original data, while the Gaussianized data approaches a standard Gaussian distribution.}
    \label{fig:iterative gaussianization}
\end{figure}

\section{Results}

\subsection{Experimental Setup}

A synthetic dataset was generated using a Gaussian Mixture Model (GMM) with a predefined number of components, weights, means, and covariance matrices. 

The diffusion models (DDPMs) used in the experiments were implemented using a fully connected neural network architecture. Each model consisted of three linear layers with ReLU activation functions. The networks were trained with varying layer widths to assess the impact of model capacity on the results. See Table \ref{tab:layer_widths} for the configurations.

To evaluate the effectiveness of the Gaussianization process, we conducted a comparative experiment using data generated from a Gaussian Mixture Model (GMM). The experiments consisted of two primary pipelines:

\begin{itemize}
    \item 
\textbf{Baseline Pipeline}  
The original data generated from the GMM was directly used to train diffusion models (DDPMs) without any preprocessing. The models aimed to learn the original data distribution and generate samples accordingly.
    \item 
\textbf{Gaussianized Data Pipeline}  
The same original data was first transformed into an approximately standard Gaussian distribution using the proposed Gaussianization method. The diffusion models (DDPMs) were then trained on the Gaussianized data. After training, the generated samples were reconstructed back to the original data distribution using the inverse Gaussianization process.
\end{itemize}

To quantitatively evaluate the performance of the Gaussianization and reconstruction process, we employ the {average log-likelihood} as the primary metric. The original data is generated from a Gaussian Mixture Model (GMM), allowing the true probability density function \( p(x) \) to be directly computed. The log-likelihood of a generated sample \( x_j \) is given by:
\begin{equation}
    \log p(x_j) = \log \left( \sum_{k=1}^{K} \pi_k \, \mathcal{N}(x_j \mid \mu_k, \Sigma_k) \right),
\end{equation}
where \( K \) is the number of Gaussian components, \( \pi_k \) represents the mixture weights, and \( \mathcal{N}(x_j \mid \mu_k, \Sigma_k) \) denotes the probability density of a multivariate normal distribution. The average log-likelihood for \( n \) generated samples is then computed as:
\begin{equation}
    \text{Average Log-Likelihood} = \frac{1}{n} \sum_{j=1}^{n} \log p(x_j).
\end{equation}
A higher average log-likelihood indicates better alignment between the generated samples \( \{x_j\}_{j=1}^n \) and the true distribution \( p(x) \). This suggests that the Gaussianization and reconstruction process successfully approximates the target distribution.

\begin{table}[t]
    \centering
    \caption{Network Width Configurations by Layer}
    \label{tab:layer_widths}
    \begin{tabular}{l c c c}
        \hline
        \textbf{Configuration} & \textbf{1st Layer} & \textbf{2nd Layer} & \textbf{3rd Layer} \\
        \hline
        \hline
        \textbf{Width 16}  & 16 & 32  & 64 \\
        \textbf{Width 32}  & 32 & 64  & 128 \\
        \textbf{Width 64}  & 64 & 128 & 256 \\
        \textbf{Width 128} & 128 & 256 & 512 \\
        \hline
    \end{tabular}
\end{table}

\subsection{Reconstruction Steps Comparison}


\begin{figure}[t]
    \centering

    \begin{subfigure}{0.9\textwidth}
        \caption{Baseline, Width 16}
        \includegraphics[width=\textwidth]{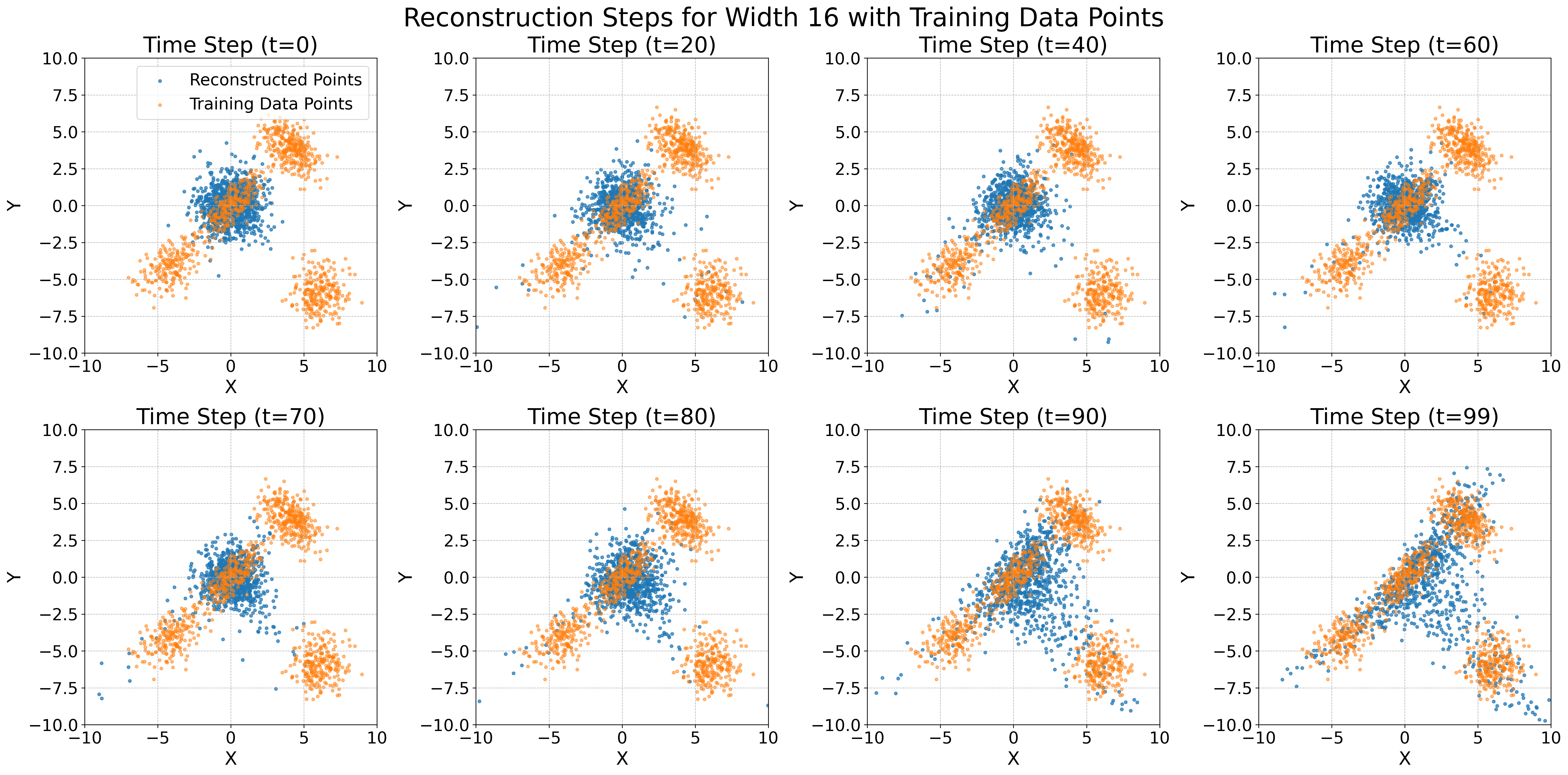}
        
    \end{subfigure}
    \hfill
    \begin{subfigure}{0.9\textwidth}
        \caption{Gaussianized, Width 16}
        \includegraphics[width=\textwidth]{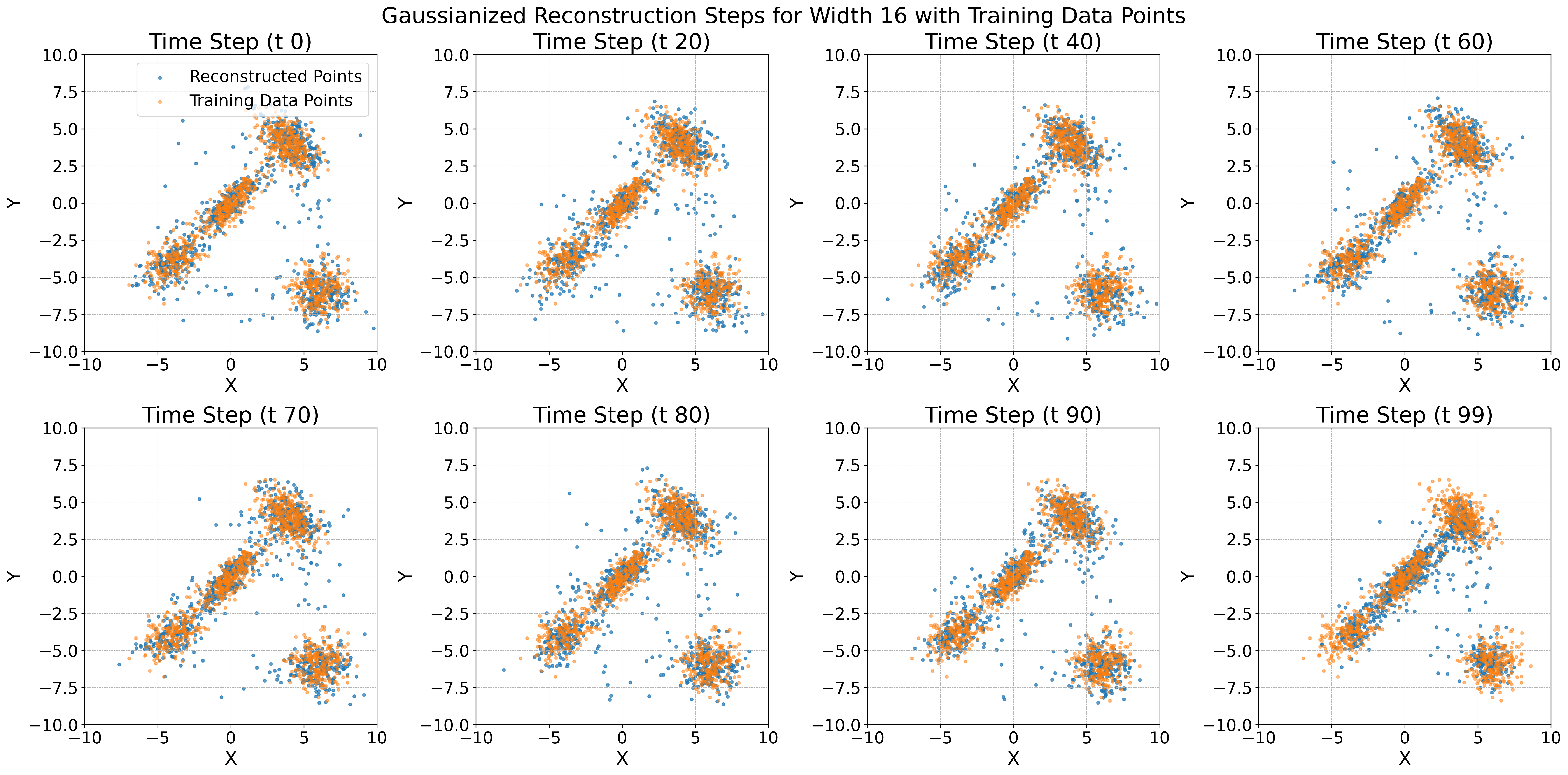}
    \end{subfigure}
    \caption{Snapshots of reconstruction steps by the diffusion model using network width 16. Orange dots represent training data sampled from the Gaussian mixture model. The blue dots are reconstruction by the diffusion model at different time points.  (a) Baseline, (b) Gaussianized.}
    \label{fig:snapshots_16}
\end{figure}

Figure \ref{fig:snapshots_16} illustrates a comparison of reconstruction steps between the baseline (original data) and Gaussianized data pipelines under network width 16 (See Figs.~\ref{fig:snapshots_32}, \ref{fig:snapshots_64}, \ref{fig:snapshots_128} for the results of widths 32, 64, 128). 

Compared to the baseline, the Gaussianized pipeline converges to the target distribution much faster. In the 100-step inference process, the baseline typically begins to approach the target distribution only after 80 steps. This suggests that, prior to the bifurcation point, the baseline spends a significant number of steps on inference without contributing meaningfully to the reconstruction of the target distribution. In contrast, the Gaussianized pipeline can approximate the target distribution at the initial time step because the most significant nonlinear dependencies is already captured by the Gaussianization preprocessing.

Further, the Gaussianized data demonstrates significantly greater stability during intermediate reconstruction steps, resulting in smooth transitions between distributions across time steps. On the other hand, the baseline pipeline performs reconstruction through bifurcation instability, leading to abrupt transitions in the reconstruction trajectory.

Finally, while the baseline pipeline can accurately reconstruct distributions at high network width, the Gaussianized pipeline can also do so at low network width. This indicates that the Gaussianization method maintains reconstruction quality while simultaneously improving efficiency and stability.

In summary, the Gaussianized pipeline consistently outperforms the baseline in terms of convergence speed and stability, while preserving reconstruction quality across different network widths.

\subsection{Log-Likelihood and Training Loss Comparison}

\begin{figure}[t!]
    \centering
    \begin{subfigure}{0.8\textwidth}
        \caption{ Baseline Log-Likelihood with Selected Inference Time Steps}
        \includegraphics[width=\textwidth]{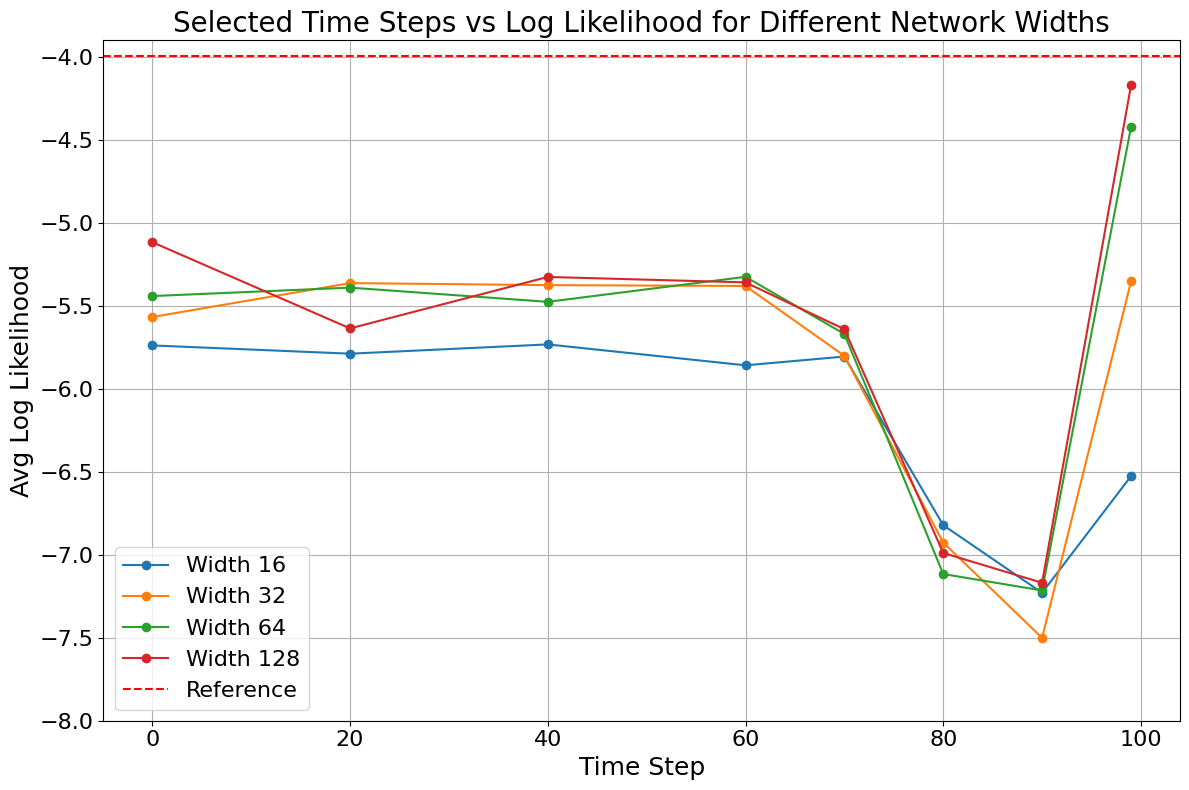}
        
    \end{subfigure}

    \hfill

    \begin{subfigure}{0.8\textwidth}
        \caption{ Gaussianized Log-Likelihood with Selected Inference Time Steps}
        \includegraphics[width=\textwidth]{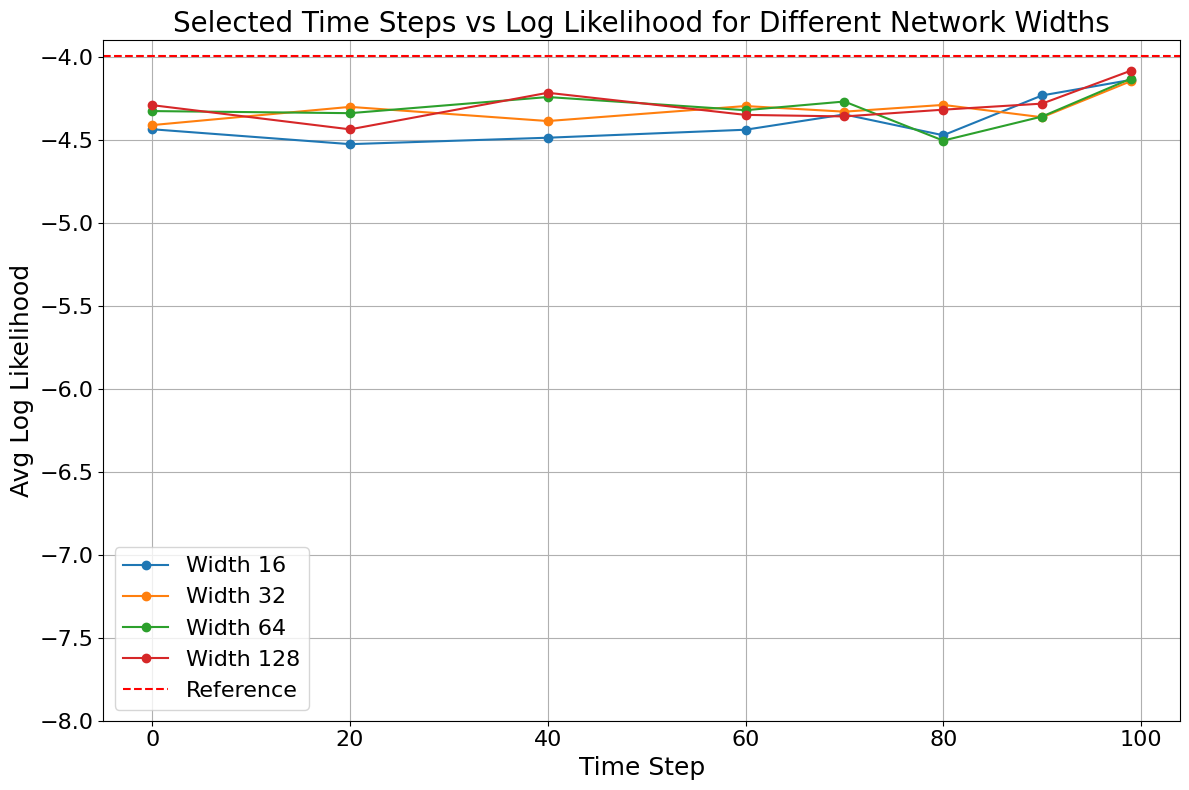}
     
    \end{subfigure}
    \caption*{Figure 5: Log-likelihood comparison for selected inference time steps. The red dashed line represents the reference average log-likelihood, which is computed from the original Gaussian mixture data using the true probability density function.}  
    \label{fig:log_likelihood_comparison}
\end{figure}

\begin{figure}[t!]
    \centering
    \begin{subfigure}{0.8\textwidth}
        \caption{ Baseline Training Loss }
        \includegraphics[width=\textwidth]{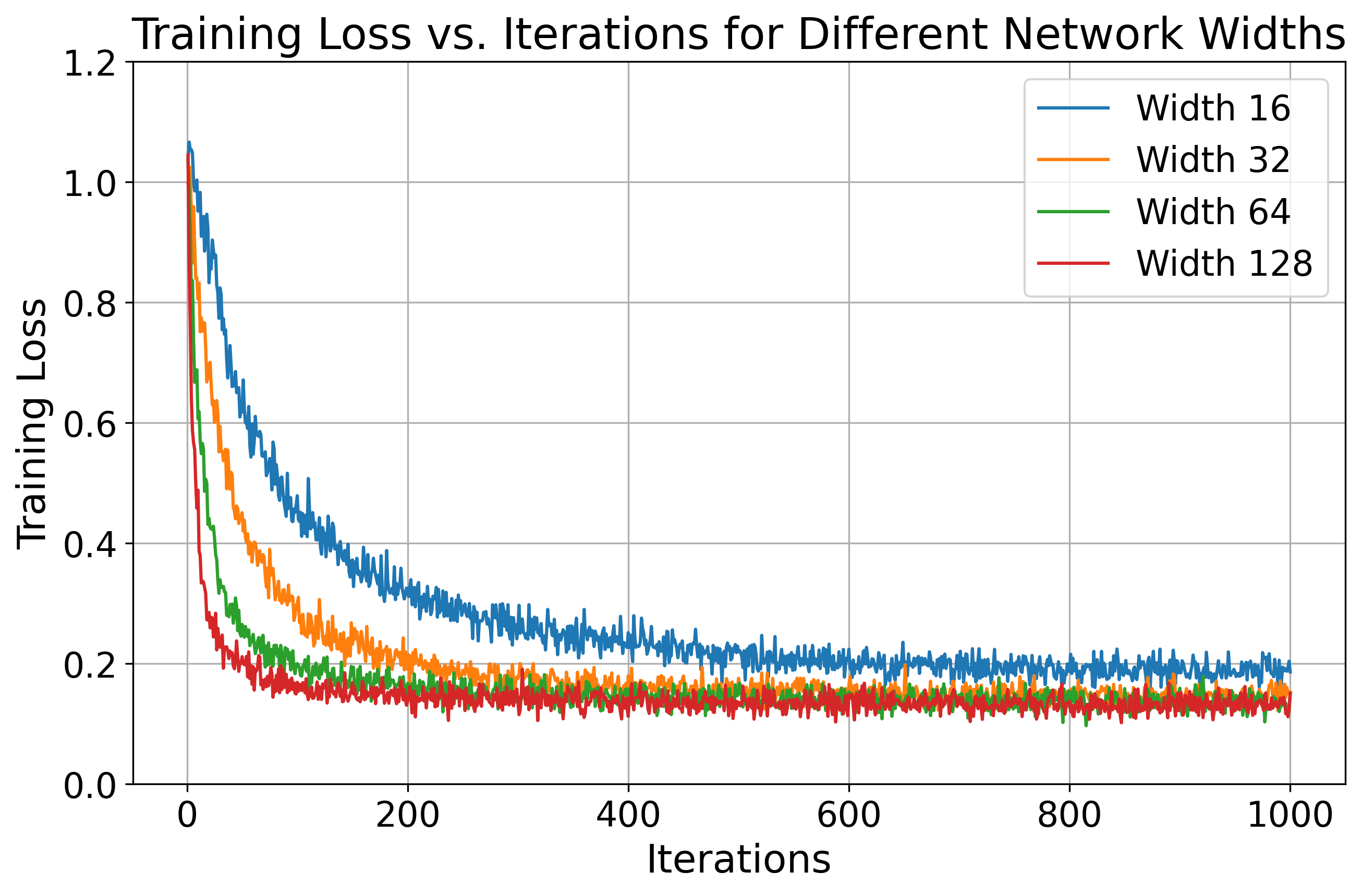}
    \end{subfigure}

    \hfill

    \begin{subfigure}{0.8\textwidth}
        \caption{ Gaussianized Training Loss }
        \includegraphics[width=\textwidth]{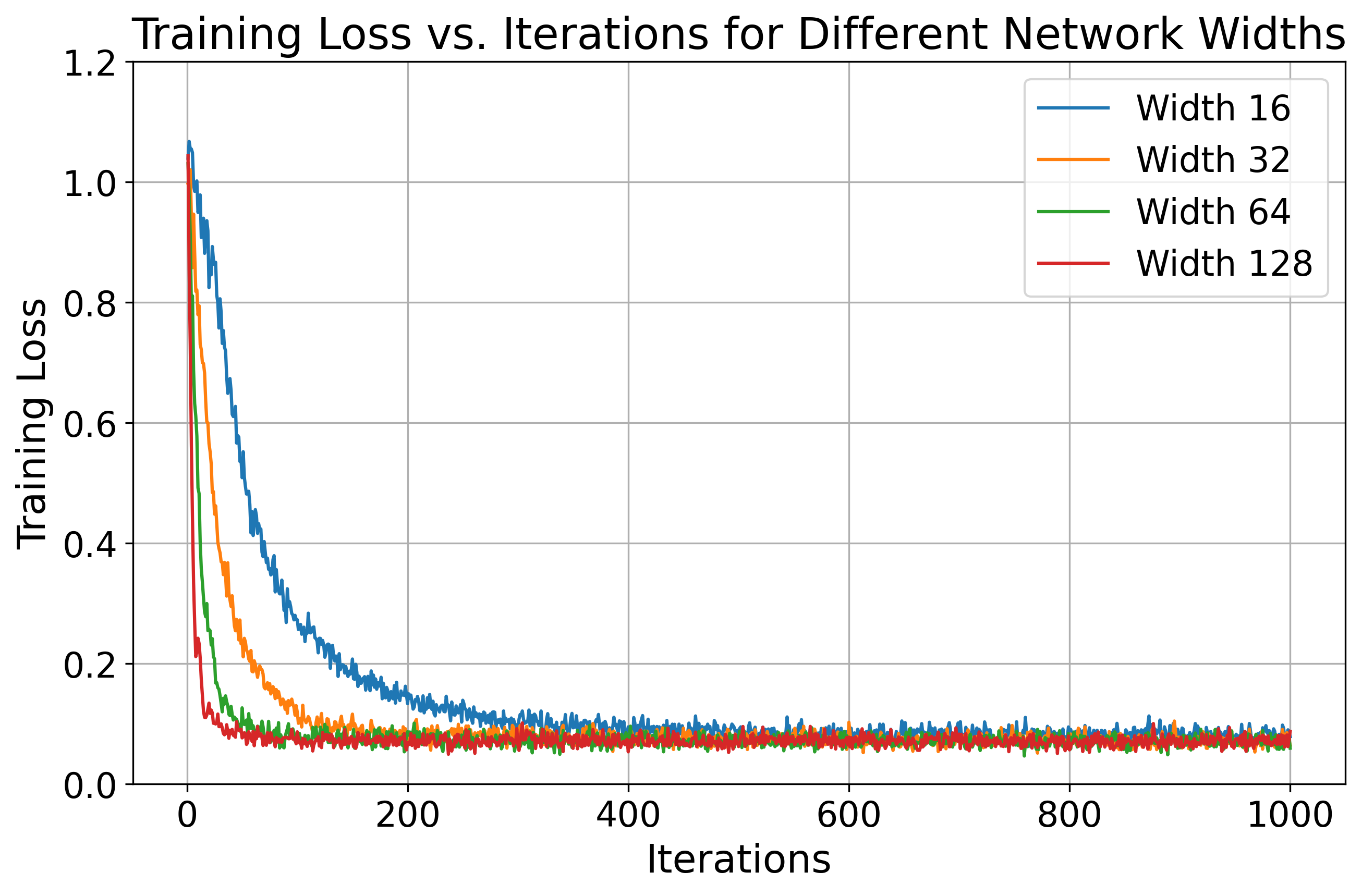}
        
    \end{subfigure}
    \caption*{Figure 6: Training loss comparison across iterations. } 
    \label{fig:training_loss_comparison}
\end{figure}

To evaluate the performance of the two pipelines, we compared the log-likelihood across time steps and the training loss curves for different widths. Figures 5 and 6 compare the log-likelihood and training loss of the baseline pipeline and the Gaussianized pipeline across different network widths, demonstrating the impact of Gaussianization preprocessing on the performance of diffusion models.


The log-likelihood in the baseline pipeline remains relatively stable during the early and intermediate stages of inference but shows a noticeable improvement only towards the later stages (e.g., after step 80) (Fig.~5(a)). This indicates that the baseline pipeline requires more time steps to converge to the target distribution, particularly after the bifurcation transition point.

In contrast, the Gaussianized pipeline achieves a higher log-likelihood value within the first 20 steps of inference and maintains it stably, demonstrating a faster convergence to the target distribution (Fig.~5(b)). The Gaussianization preprocessing effectively transforms the data distribution to better align with a Gaussian form, reducing unnecessary inference steps. This improvement is particularly pronounced in narrower networks (e.g., with a width of 16), where the log-likelihood values of the Gaussianized pipeline consistently surpass those of the baseline, including at the final reconstruction time step of $t = 100$.


We now turn to the train loss. The training loss in the baseline pipeline decreases slowly during the initial stages and gradually converges as the number of iterations increases. Narrower networks exhibit relatively slower convergence rates (Fig.~6(a)). In constrast, the Gaussianized pipeline achieves slightly faster reductions in training loss and converges with fewer iterations (Fig.~6(b)). This suggests that simplifying the training data distribution via Gaussianization marginally accelerates the optimization process. Additionally, the Gaussianized pipeline plateaus at a lower final training loss.

In summary, compared to the baseline pipeline, the Gaussianized pipeline demonstrates a certain efficiency advantage during inference. Specifically, the Gaussianized pipeline converges to the target distribution within the first 20 steps and consistently maintains higher log-likelihood values, whereas the baseline pipeline typically requires over 80 steps to achieve similar performance. This efficiency improvement is particularly pronounced in narrower networks, such as those with a width of 16.

Further, during the training phase, the Gaussianized pipeline accelerates the convergence of training loss by simplifying the data distribution. Compared to the baseline pipeline, the Gaussianized method achieves the same loss value with fewer iterations.

Despite transforming the data distribution, the Gaussianized pipeline achieves low training losses and final-step log-likelihood values that are nearly identical to those of the target distribution, regardless of network width. This indicates that the Gaussianization method improves both training and inference efficiency without compromising the quality of the generated samples, successfully maintaining robust reconstruction of the target distribution.

\section{Discussion}

In this study, we proposed a novel Gaussianization preprocessing method that transforms the training data distribution to enhance the performance of diffusion models. Based on experimental results and theoretical analysis, we summarize the key points of discussion as follows:

The experimental results demonstrate that Gaussianization preprocessing significantly improves the inference efficiency of diffusion models. In the baseline pipeline, a substantial portion of inference steps is wasted in ineffective regions of the distribution, particularly before the bifurcation point. In contrast, the Gaussianized pipeline remains close to the target distribution in the early stages of inference, thereby substantially reducing the number of required inference steps. This not only accelerates the generation process but also reduces computational resource consumption. Notably, the efficiency improvement is especially pronounced for narrower networks, highlighting the potential applicability of Gaussianization preprocessing in resource-constrained scenarios. By simplifying the data distribution, Gaussianization preprocessing reduces the likelihood of bifurcation points during inference, making the effects of early time steps on the generation outcomes less significant. Furthermore, the Gaussianized data distribution aligns more closely with the core assumptions of the Gaussian late start strategy, where pre-bifurcation trajectories are simplified into forms resembling Gaussian distributions. This alignment enhances the reliability and effectiveness of the Gaussian late start strategy.

Analysis of training loss dynamics indicates that Gaussianization preprocessing accelerates the optimization process to some extent. By transforming complex high-dimensional data distributions into forms that approximate independent Gaussian distributions, Gaussianized pipelines simplify the training objective, enabling the model to capture the global characteristics of the target distribution more efficiently. This contributes to more effective and faster training of diffusion models.

Although Gaussianization preprocessing alters the data distribution, the final quality of samples generated by the Gaussianized pipeline remains almost identical to that of the baseline pipeline. This demonstrates that Gaussianization preprocessing not only enhances efficiency but also largely retains the model's ability to learn the target distribution. These results theoretically validate the importance of optimizing data distributions in improving generative model performance and provide new directions for further enhancing diffusion models.

Despite its notable advantages in efficiency and stability, Gaussianization preprocessing has certain limitations in practical applications. For instance, the current method relies heavily on Independent Component Analysis (ICA) and Kernel Density Estimation (KDE), which perform well on low-dimensional data but may struggle to efficiently handle high-dimensional data. Additionally, in scenarios involving high-dimensional data or complex distributions, the preprocessing process may introduce additional computational overhead, posing challenges for scalability to high-resolution image data. To mitigate computational costs, Gaussianization may be selectively applied to the leading ICA components, as these components tend to exhibit the highest non-Gaussianity. By focusing on the most non-Gaussian components, the method can retain its benefits while improving efficiency and scalability.

Future work could explore more efficient preprocessing techniques, such as deep learning-based adaptive Gaussianization networks, to enhance the flexibility and scalability of the Gaussianization process. Furthermore, integrating Gaussianization with other advanced generative model techniques, such as conditional generative networks, could further improve its applicability. Additionally, task-specific Gaussianization strategies, such as those designed for text-to-image generation or super-resolution reconstruction, could be developed to further optimize the performance and quality of the generation process. These advances could significantly broaden the applicability of Gaussianization preprocessing in practical scenarios.

Gaussianization preprocessing provides an efficient optimization strategy for diffusion models, demonstrating significant advantages in inference efficiency, training acceleration, and stability, while maintaining high-quality sample generation. With further optimization and exploration, Gaussianization methods hold great promise for supporting the development of high-performance diffusion models in practical applications.


\newpage

\makeatletter
\g@addto@macro\appendix{%
  \setcounter{figure}{0}%
  \setcounter{table}{0}%
  \setcounter{equation}{0}%
  \renewcommand{\thefigure}{S\arabic{figure}}%
  \renewcommand{\thetable}{S\arabic{table}}%
  \renewcommand{\theequation}{S\thesection.\arabic{equation}}%
  \def\fnum@figure{\textbf{Fig.\thefigure}}%
  \def\fnum@table{\textbf{Supplementary \thetable}}%
}
\makeatother
\appendix

\begin{figure}[H]
    \centering
    \begin{subfigure}{0.9\textwidth}
        \caption{Baseline, Width 32}
        \includegraphics[width=\textwidth]{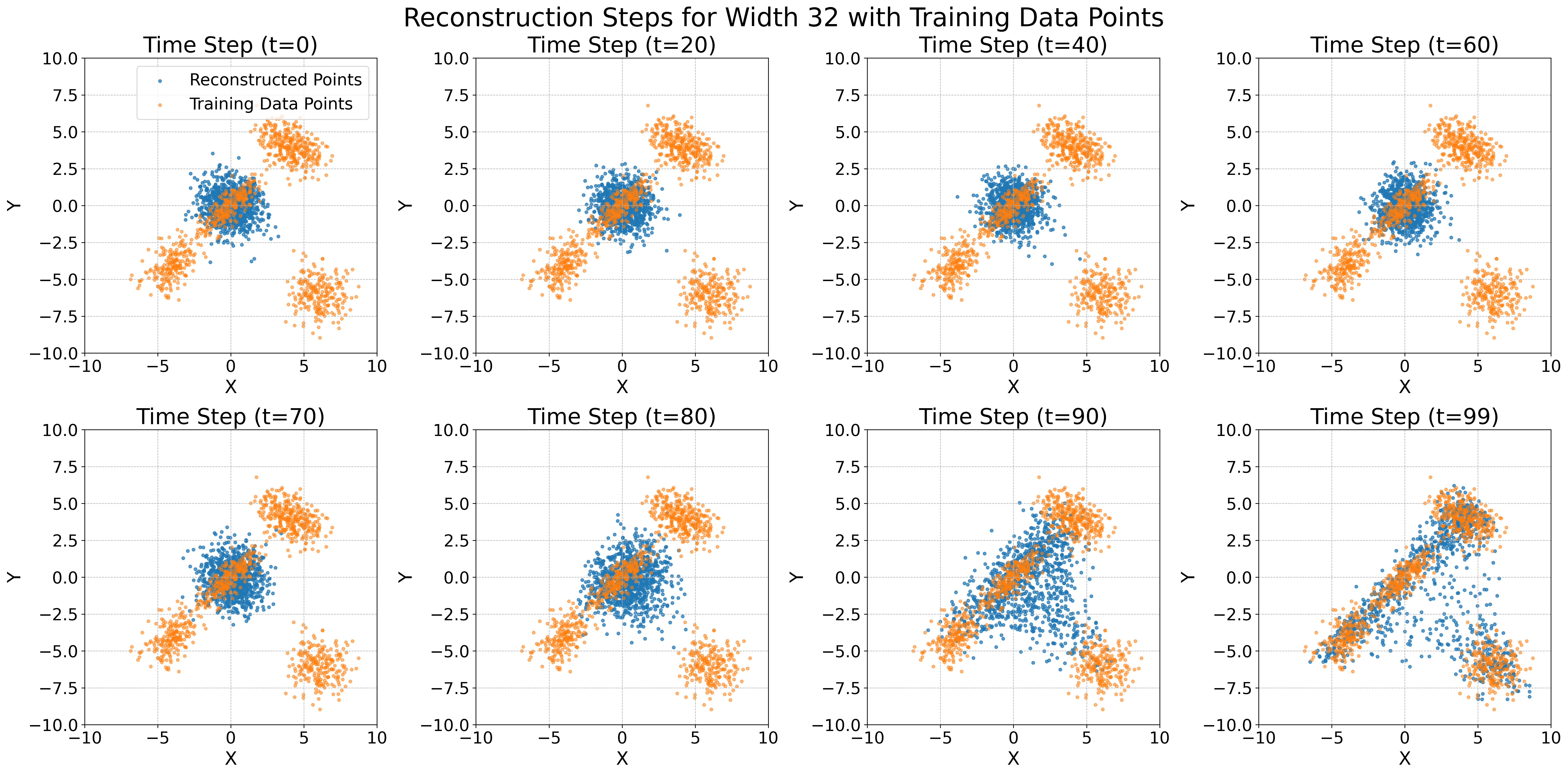}
    \end{subfigure}
    \hfill
    \begin{subfigure}{0.9\textwidth}
        \caption{Gaussianized, Width 32}
        \includegraphics[width=\textwidth]{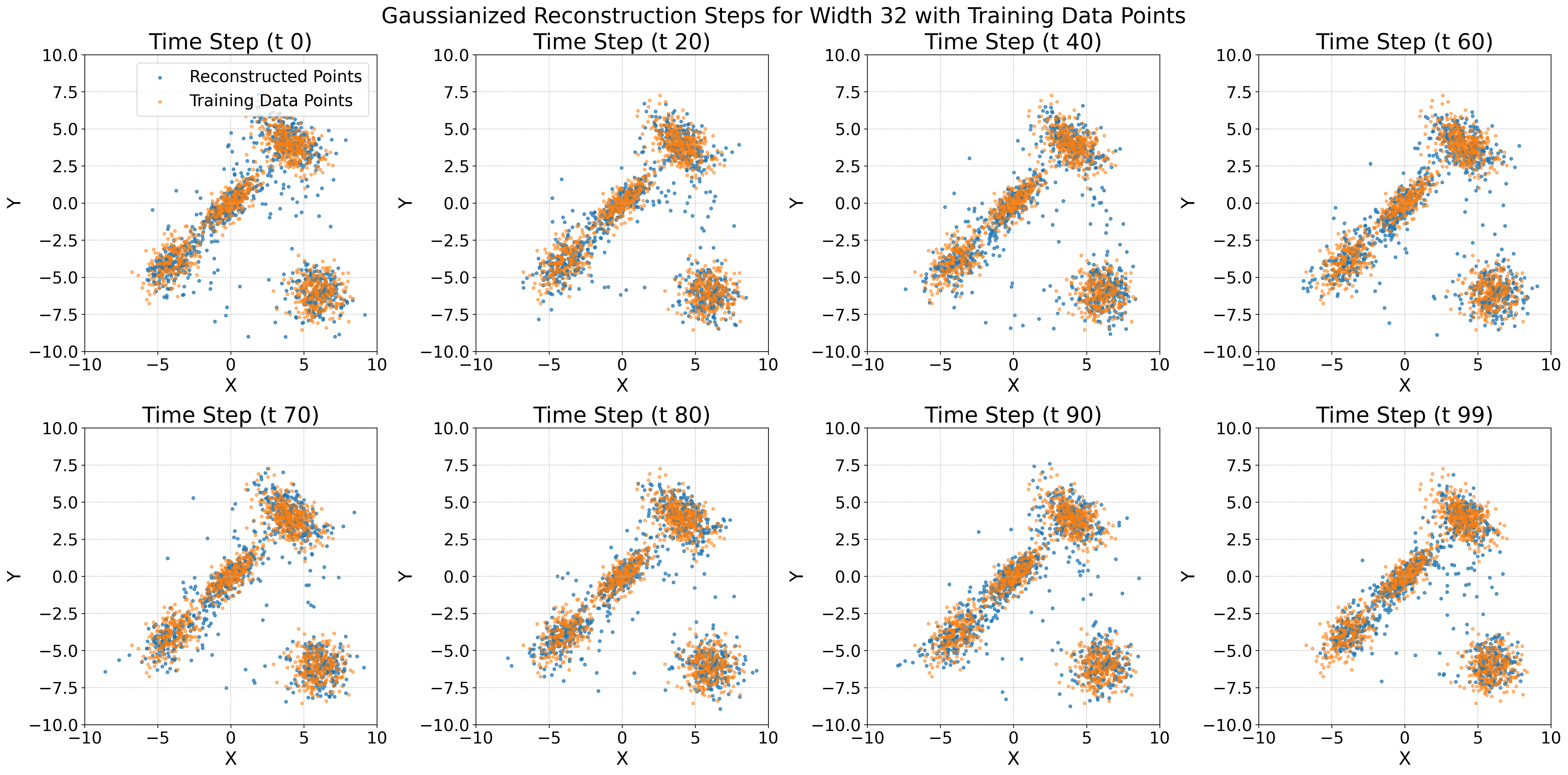}
    \end{subfigure}
    \caption{Snapshots of reconstruction steps by the diffusion model using the network width 32. (a) Baseline, (b) Gaussianized. Presentation follows Fig.~\ref{fig:snapshots_16}}.
    \label{fig:snapshots_32}
\end{figure}

\begin{figure}[H]
    \centering
    \begin{subfigure}{0.9\textwidth}
        \caption{Baseline, Width 64}
        \includegraphics[width=\textwidth]{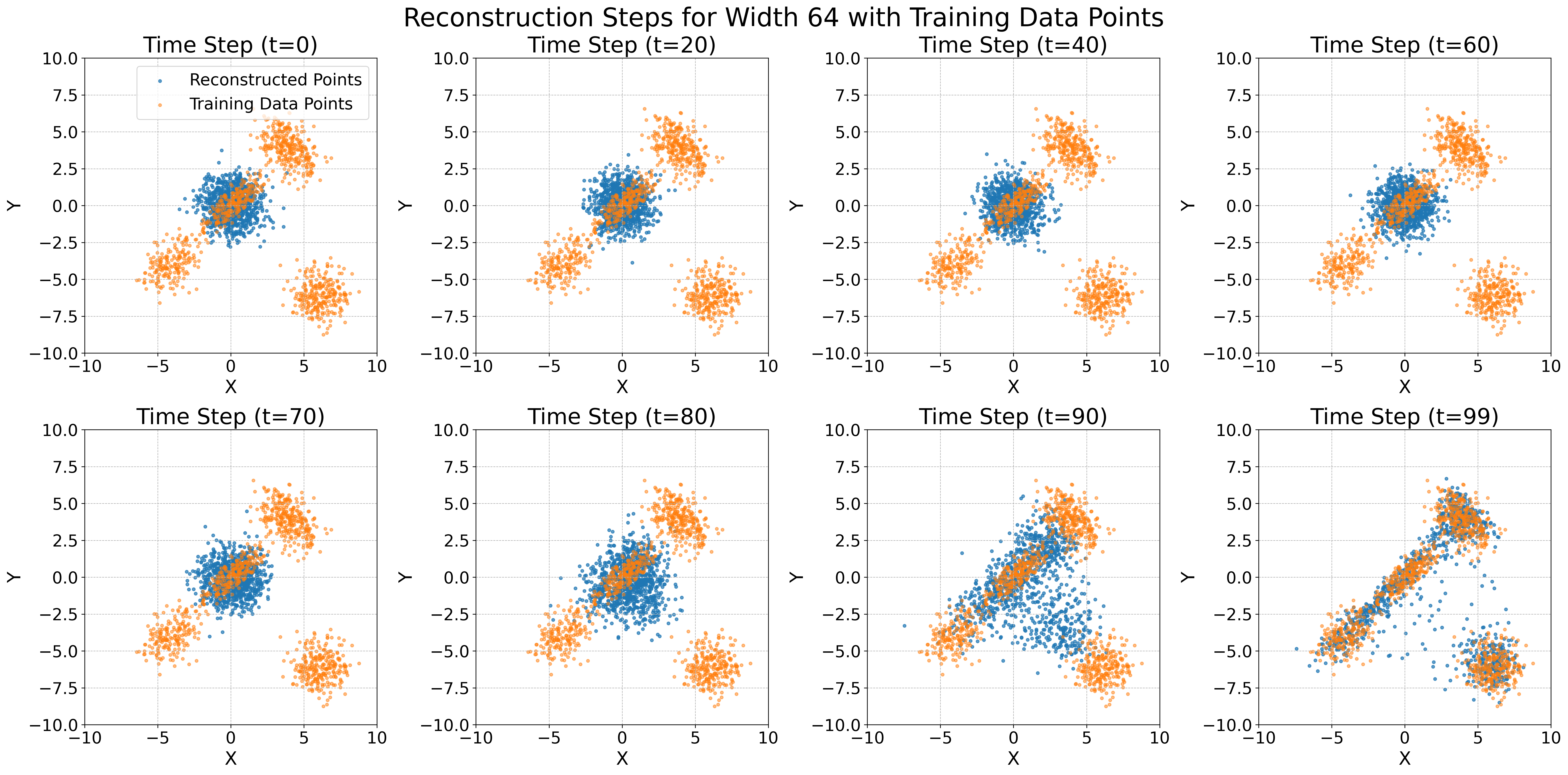}
    \end{subfigure}
    \hfill
    \begin{subfigure}{0.9\textwidth}
        \caption{Gaussianized, Width 64}
        \includegraphics[width=\textwidth]{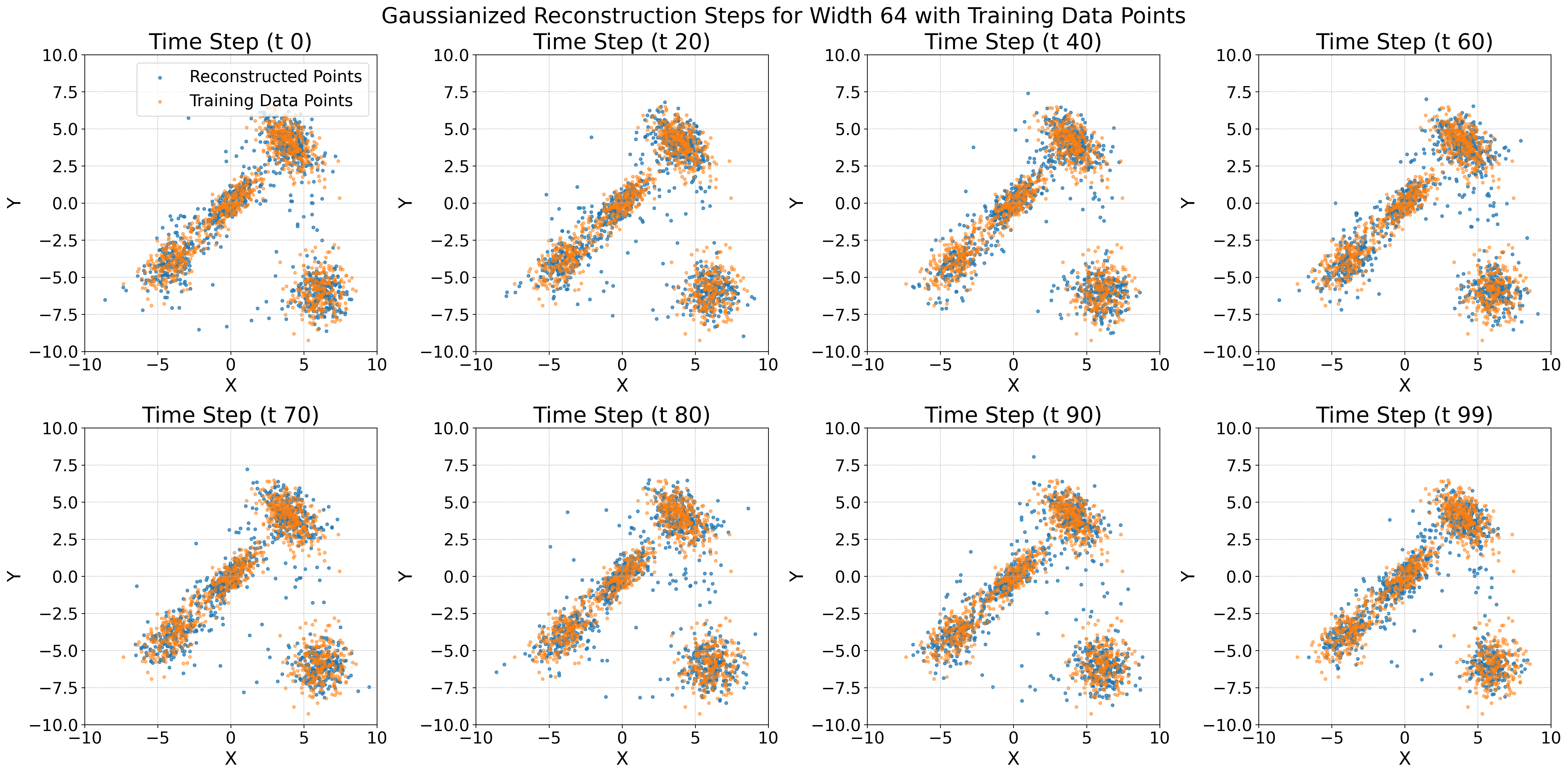}
    \end{subfigure}
    \caption{Snapshots of reconstruction steps by the diffusion model using the network width 64. (a) Baseline, (b) Gaussianized. Presentation follows Fig.~\ref{fig:snapshots_16}}.
    \label{fig:snapshots_64}
\end{figure}

\begin{figure}[H]
    \centering
    \begin{subfigure}{0.9\textwidth}
        \caption{Baseline, Width 128}
        \includegraphics[width=\textwidth]{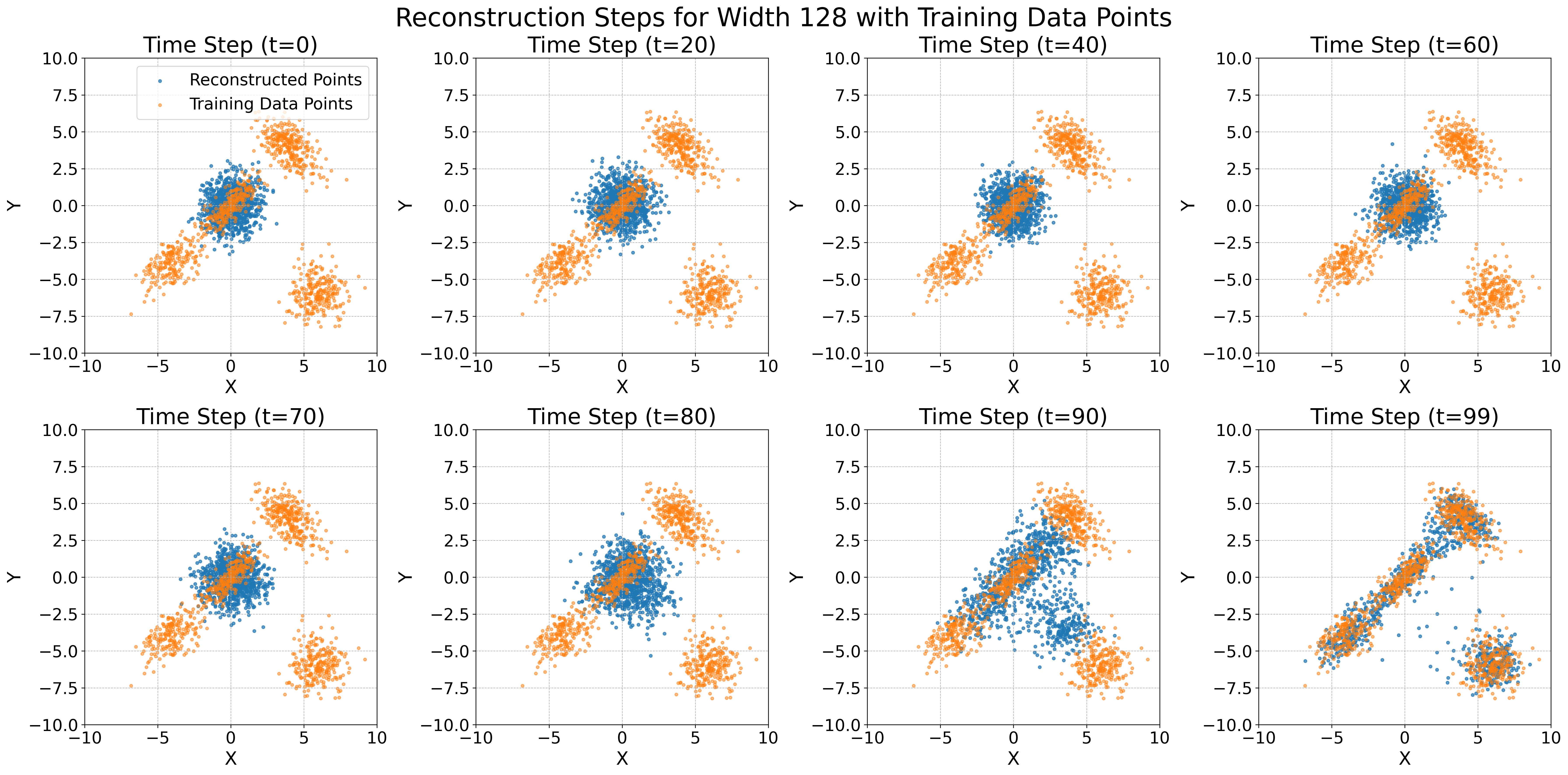}
    \end{subfigure}
    \hfill
    \begin{subfigure}{0.9\textwidth}
        \caption{Gaussianized, Width 128}
        \includegraphics[width=\textwidth]{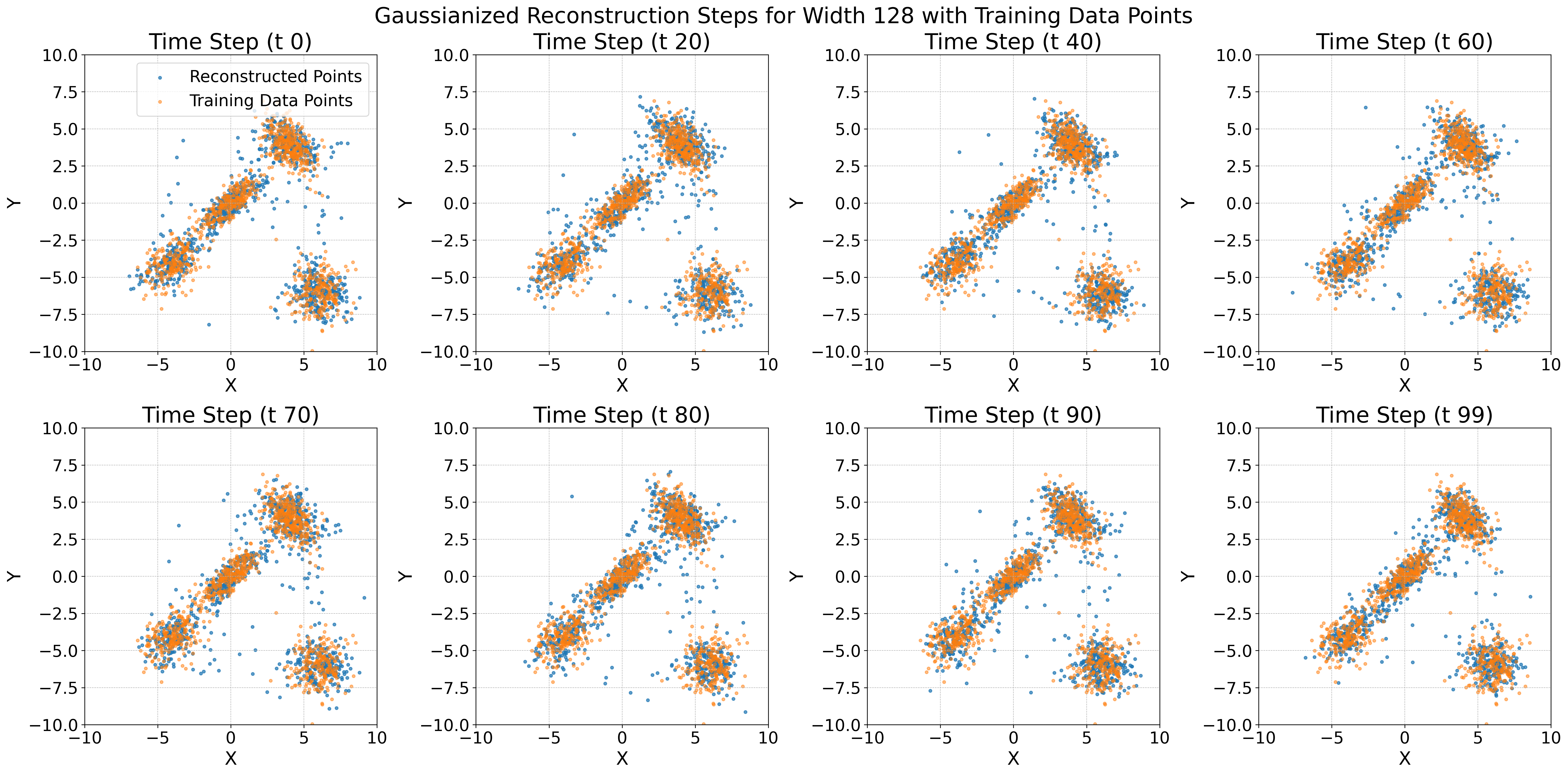}
    \end{subfigure}
    \caption{Snapshots of reconstruction steps by the diffusion model using the network width 128. (a) Baseline, (b) Gaussianized. Presentation follows Fig.~\ref{fig:snapshots_16}}.
    \label{fig:snapshots_128}
\end{figure}

\end{document}